\documentclass[a4paper,fleqn]{cas-sc}

\usepackage[authoryear,longnamesfirst]{natbib}
\usepackage{graphicx}
\usepackage{amsmath}
\usepackage{booktabs}
\usepackage{multirow}
\usepackage{graphicx}
\usepackage{subfigure}
\usepackage{algorithm}  
\usepackage{algpseudocode}

\def\tsc#1{\csdef{#1}{\textsc{\lowercase{#1}}\xspace}}
\tsc{WGM}
\tsc{QE}
\tsc{EP}
\tsc{PMS}
\tsc{BEC}
\tsc{DE}

\begin{document}
\let\WriteBookmarks\relax
\def\floatpagepagefraction{1}
\def\textpagefraction{.001}
\shorttitle{Bidirectional Loss Function for Label Enhancement and Distribution Learning}
\shortauthors{Xinyuan Liu et~al.}

\title [mode = title]{Bidirectional Loss Function for Label Enhancement and Distribution Learning}      
\tnotemark[1]

\tnotetext[1]{This work is supported by the National Natural Science Foundation of China under Grant No. 61573273.}

\author[mymainaddress]{Xinyuan Liu}[style=chinese]
 
 \author[mymainaddress]{Jihua Zhu}[style=chinese, orcid=0000-0002-3081-8781]
 \cormark[1]
 
 \author[mymainaddress]{Qinghai Zheng}[style=chinese]
 
 \author[mymainaddress]{Zhongyu Li}[style=chinese]
 
 \author[mymainaddress]{Ruixin Liu}[style=chinese]
 
 \author[mysecondaddress]{Jun Wang}[style=chinese]
 
 \address[mymainaddress]{Lab of VCAML, School of Software Engineering, Xi'an Jiaotong University, Xi'an 710049, People's Republic of China}
 
 \address[mysecondaddress]{School of Communication and Information Engineering, Shanghai University, Shanghai 200072, People's Republic of China}
 
 \cortext[cor1]{Corresponding author, email: zhujh@xjtu.edu.cn.}

\begin{abstract}
Label distribution learning (LDL) is an interpretable and general learning paradigm that has been applied in many real-world applications. In contrast to the simple logical vector in single-label learning (SLL) and multi-label learning (MLL), LDL assigns labels with a description degree to each instance. In practice, two challenges exist in LDL, namely, how to address the dimensional gap problem during the learning process of LDL and how to exactly recover label distributions from existing logical labels, i.e., Label Enhancement (LE). For most existing LDL and LE algorithms, the fact that the dimension of the input matrix is much higher than that of the output one is alway ignored and it typically leads to the dimensional reduction owing to the unidirectional projection. The valuable information hidden in the feature space is lost during the mapping process. To this end, this study considers bidirectional projections function which can be applied in LE and LDL problems simultaneously. More specifically, this novel loss function not only considers the mapping errors generated from the projection of the input space into the output one but also accounts for the reconstruction errors generated from the projection of the output space back to the input one. This loss function aims to potentially reconstruct the input data from the output data. Therefore, it is expected to obtain more accurate results. Finally, experiments on several real-world datasets are carried out to demonstrate the superiority of the proposed method for both LE and LDL.
\end{abstract}


\maketitle

\section{Introduction}
Learning with ambiguity has become one of the most prevalent research topics. The traditional way to solve machine learning problems is based on single-label learning (SLL) and multi-label learning (MLL) \cite{tsoumakas2007multiMLL,xu2016multi}. Concerning the SLL framework, an instance is always assigned to one single label, whereas in MLL an instance may be associated with several labels. The existing learning paradigms of SLL and MLL are mostly based on the so-called \textit{problem transformation}. However, neither SLL nor MLL address the problem stated as ``at which degree can a label describe its corresponding instance,'' i.e., the labels have different importance on the description of the instance. It is more appropriate for the importance among candidate labels to be different rather than exactly equal. Taking the above problem into account, a novel learning paradigm called label distribution learning (LDL) \cite{geng2013LDL} is proposed. Compared with SLL and MLL, LDL labels an instance with a real-valued vector that consists of the description degree of every possible label to the current instance. Detail comparison is visualized in Fig. \ref{fig:Comparison among SLL, MLL and LDL}. Actually, LDL can be regarded as a more comprehensive form of MLL and SLL. However, the tagged training sets required by LDL are extremely scarce owing to the heavy burden of manual annotation. Considering the fact that it is difficult to directly attain the annotated label distribution, a process called label enhancement (LE) \cite{xu2018LE} is also proposed to recover the label distributions from logical labels. Taking LE algorithm, the logical label $l \in\{0,1\}^{\mathrm{c}}$ of conventional MLL dataset can be recovered into the label distribution vector by mining the topological information of input space and label correlation \cite{he2019joint}. 
\begin{figure}[]
	\centering
	\includegraphics[width=1\columnwidth]{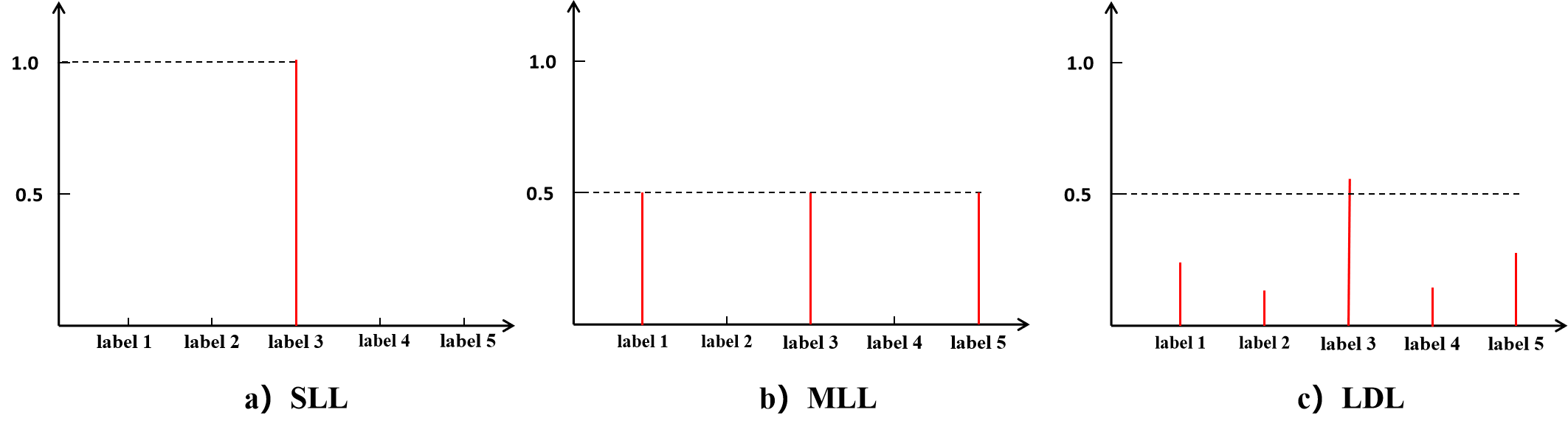}
	\caption{Visualized comparison among SLL, MLL and LDL}
	\label{fig:Comparison among SLL, MLL and LDL}
\end{figure}

Many relevant algorithms of LDL and LE have been proposed in recent years. These algorithms have progressively boosted the performance of specific tasks. For instance, LDL is widely applied in facial age estimation application. Geng et al. \cite{geng2014facialadaptive} proposed a specialized LDL framework that combines the maximum entropy model \cite{berger1996maximum} with IIS optimization, namely IIS-LDL. This approach not only achieves better performance than other traditional machine learning algorithms but also becomes the foundation of the LDL framework. In other works, Yang et al.’s \cite{fan2017labelreNetFacial} attempt to take into account both facial geometric and convolutional features resulted in remarkably improving efficiency and accuracy. As mentioned above, the difficulty of acquiring labeled datasets restricts the development of LDL algorithms. After presenting several LE algorithms, Xu et al. \cite{XuLv-14} adapted LDL into partial label learning (PLL) with recovered label distributions via LE. Although these methods have achieved significant performance, one potential problem yet to be solved is that they suffer from the discriminative information loss problem, which is caused by the dimensional gap between the input data matrix and the output one. Importantly, it is entirely possible that these existing methods miss the essential information that should be inherited from the original input space, thereby degrading the performance. 

As discussed above, the critical point of previous works on LDL and LE is to establish a suitable loss function to fit label distribution data. In previous works, only a unidirectional projection $\mathcal{X} \mapsto \mathcal{Y}$ between input and output space is learned. In this paper, we present a bi-directional loss function with a comprehensive reconstruction constraint. Such function can be applied in both LDL and LE to maintain the latent information. Inspired by the auto-encoder paradigm \cite{kodirov2017SAE,cheng2019multi}, our proposed method builds the reconstruction projection $\mathcal{Y} \mapsto \mathcal{X}$ with the mapping projection to preserve the otherwise lost information. More precisely, optimizing the original loss is the \textit{mapping} step, while minimizing the \textit{reconstruction} error is the reconstruction step. In contrast to previous loss functions, the proposed loss function aims to potentially reconstruct the input data from the output data. Therefore, it is expected to obtain more accurate results than other related loss functions for both LE and LDL problems. Adequate experiments on several well-known datasets demonstrate that the proposed loss function achieves superior performance.

The main contributions of this work are delivered as: 
\begin{enumerate}[1)]
\item the reconstruction projection from label space to instance space is considered for the first time in the LDL and LE paradigms;
\item a bi-directional loss function that combines mapping error and reconstruction error is proposed;
\item the proposed method can be used not only in LDL but also for LE. 
\end{enumerate}

We organize the rest of this paper as follows. Firstly, related work about LDL and LE methods is reviewed in Section 2. Secondly, the formulation of LE as well LDL and the proposed methods, i.e., BD-LE and BD-LDL are introduced in Section 3. After that, the results of comparison experiment and ablation one are shown in Section 4. The influence of parameters is also discussed in Section 4. Finally, conclusions and future work exploration are summarized in section 5.

\section{Related work}
In this section, we briefly summarize the related work about LDL and LE methods.
\subsection{Label Distribution Learning}

The proposed LDL methods mainly focus on three aspects, namely model assumption, loss function, and the optimization algorithm. The maximum entropy model \cite{berger1996maximum} is widely used to represent the label distribution in the LDL paradigm\cite{xu2019latent,ren2019label}. Maximum entropy model naturally agrees with the character of description degree in LDL model. However, such an exponential model is sometimes not comprehensive enough to accomplish a complex distribution. To overcome this issue, Gent et al. \cite{xing2016logisticBoosting} proposed a LDL family based on a boosting algorithm to extend the traditional LDL model. Inspired by the M-SVR algorithm, LDVSR \cite{geng2015preMovieOpnion} is designed for the movie 
opinion prediction task. Furthermore, CPNN \cite{geng2013facialestimation} combines a neural network with the LDL paradigm to improve the effectiveness of facial age estimation applications. What's more, recent work \cite{ren2019specific,xu2017incompleteLDL} has proved that linear model is also able to achieve a relatively strong representation ability and a satisfying result. As reviewed above, most existing methods build the mapping from feature space to label space in an unidirectional way so that it is appropriate to take the bi-directional constraint into consideration. 

Concerning the loss function, LDL aims at learning the model to predict unseen instances’ distributions which are similar to the true ones. The criteria to measure the distance between two distributions, such as the Kullback–Leibler (K-L) divergence, is always chosen as the loss function \cite{jia2018labelCorrelation,geng2010facialestimation}. Owing to the asymmetry of the K-L divergence, Jeffery’s divergence is used in xxx \cite{zhou2015emotion} to build LDL model for facial emotion recognition. For the sake of easier computation, it is reasonable to adopt the Euclidean distance in a variety of tasks, e.g., facial emotion recognition \cite{jia2019facialEDLLRL}. 

Regarding the optimization method, SA-IIS \cite{geng2016LDL} utilizes the improved iterative scaling (IIS) method whose performance is always worse \cite{Maxinum_entropy} than the other optimization. Fortunately, by leveraging the L-BFGS \cite{nocedal2006numericalOPtimization} optimization method, we maintain the balance between efficiency and accuracy, especially in SA-BFGS \cite{geng2016LDL} and EDL \cite{zhou2015emotion}. 
With the complexity of proposed model greater, the number of parameters to be optimized is more than one. Therefore, it is more appropriate to introduce the alternating direction method of multipliers (ADMM) \cite{boyd2011distributedADMM} when the loss function incorporates additional inequality and equality constraints. 
In addition, exploiting the correlation among labels or samples can increasingly boost the performance of LDL model. Jia et al. \cite{jia2018labelCorrelation} proposed LDLLC to take the global label correlation into account with introducing the Person's correlation between labels. It is pointed out in LDL-SCL \cite{zheng2018labelCorrelationSample} and EDL-LRL \cite{jia2019facialEDLLRL} that some correlations among labels (or samples) only exist in a set of instances, which are so-called the local correlation exploration. Intuitively, the instances in the same group after clustering share the same local correlation.

What's more, it is common that the labeled data are incomplete and contaminated \cite{ma2017multi}. For the former condition, Xu et al. \cite{xu2017incompleteLDL} put forward IncomLDL-a and IncomLDL-p on the assumption that the recovered complete label distribution matrix is low-rank.  Proximal Gradient Descend (PGD) and ADMM are used for the optimization of two methods respectively. The time complexity of the first one is $O\left(1 / T^{2}\right)$, and the last one is $O\left(1 / T\right)$ but good at the accuracy. Jia et al. \cite{jia2019weakly} proposed WSLDL-MCSC which is based on the matrix completion and the exploration of samples' relevance in a transductive way when the data is under weak-supervision.

\subsection{Label Enhancement Learning}
To the best of our knowledge, there are a few researches whose topics focus on the label enhancement learning \cite{xu2018LE}. Five effective strategies have been devised during the present study. Four of them are adaptive algorithms. As discussed in \cite{geng2016LDL}, the concept of \textit{membership} used in fuzzy clustering \cite{jiang2006fuzzySVM} is similar to label distribution. Although they indicated two distinguishing semantics, they are both in numerical format. Thus, FCM \cite{el2006studyKNN} extend the calculation of membership which is used in fuzzy C-means clustering \cite{melin2005hybrid} to recover the label distribution. LE algorithm based on kernel
method (KM) \cite{jiang2006fuzzySVM} utilizes the kernel function to project the instances from origin space into a high-dimensional one. The instances are separated into two parts according to whether the corresponding logical label is 1 or not for every candidate label. Then the label distribution term, i.e. description degree can be calculated based on the distance between the instances and the center of groups. Label propagation technique \cite{wang2007labelpropagation} is used in the LP method to update the label distribution matrix iteratively with a fully-connected graph built. Since the message between samples is shared and passed on the basis of the connection graph, the logical label can be enhanced into distribution-level label. LE method adapted from manifold learning (ML) \cite{hou2016manifold} take the topological consistency between feature space and label space into consideration to obtain the recovered label distribution. The last novel strategy called GLLE \cite{xu2019labelTKDE} is specialized by leveraging the topological information of the input space and the correlation among labels. Meanwhile, the local label correlation is captured via clustering \cite{zhou2012multi}.

\section{Proposed Method}
Let $\mathcal{X}=\mathbb{R}^{m}$ denote the $m$-dimensional input space and $\mathcal{Y}=\left\{y_{1}, y_{2}, \cdots, y_{c}\right\}$ represent the complete set of labels where $c$ is the number of all possible labels. For each instance $x_{i} \in \mathcal{X}$, a simple logical label vector $l_{i} \in\{0,1\}^{c}$ is leveraged to represent which labels can describe the instance correctly. Specially, for the LDL paradigm, instance $x_{i}$ is assigned with distribution-level vector  $d_{i} $
\begin{table}[]
	\centering
	\caption{Summary of some notations}\smallskip
	\resizebox{0.5\columnwidth}{!}{
		\begin{tabular}{ll}
            \toprule
            Notations & Description \\
            \midrule
            $n$ & the number of instances \\
            $c$ & number of labels \\
            $m$ & dimension of samples \\
            $X$ & instance feature matrix \\
            $L$  & logical label matrix \\
            $D$ & label distribution matrix \\
            $\hat{W} $ &  Mapping parameter of BD-LE\\
            $\tilde{W}$ & Reconstruction parameter of BD-LE \\
            $\theta $ &  Mapping parameter of BD-LDL\\
            $\tilde{\theta}$ & Reconstruction parameter of BD-LDL \\
            \bottomrule
		\end{tabular}
	}
	\label{tab:Notations}
\end{table}

\subsection{Bi-directional for Label Enhancement}
Given a dataset $E=\left\{\left(x_{1}, l_{1}\right)\left(x_{2}, l_{2}\right), \cdots,\left(x_{n}, l_{n}\right)\right\}$, $X=\left[x _{1}, x_{2}, x_{3}, \ldots, x_{n}\right]$ and $L=\left[l_{1}, l_{2}, l_{3}, \dots, l_{n}\right]$ is defined as input matrix  and logical label matrix, respectively. According to previous discussion, the goal of LE is to transform $L$ into the label distribution matrix $D=\left[d_{1}, d_{2}, d_{3}, \dots, d_{n}\right]$.

Firstly, a nonlinear function $\varphi(\cdot)$, i.e., kernel function is defined to transform each instance ${x_i}$ into a higher dimensional feature $\varphi(x_{i})$, which can be utilized to construct the vector $\phi_{i}=\left[\varphi\left(x_{i}\right) ; 1\right]$ of corresponding instance. For each instance, an appropriate mapping parameter $\hat{\mathrm{W}}$ is required to transform the input feature $\phi_{i}$ into the label distribution $d_i$. As there is a large dimension gap between input space and output space, a lot of information may be lost during the mapping process. To address this issue, it is reasonable to introduce the parameter $\tilde{W}$ for the reconstruction of the input data from the output data. Accordingly, the objective function of LE is formulated as follows:
\begin{equation}
\label{eq:obj_le_1}
\min _{\hat{W}, \tilde{W}} L(\hat{W})+\alpha R(\tilde{W})+\frac{1}{2} \lambda \Omega(\hat{W})+\frac{1}{2} \lambda \Omega(\tilde{W})
\end{equation}
where $L$ denotes the loss function of data mapping, $R$ indicates the loss function of data reconstruction, $\Omega$ is the regularization term, $\lambda$ and $\alpha$ are two trade-off parameters. It should be noted that the LE algorithm is regarded as a pre-processing of LDL methods and it dose not suffer from the over-fitting problem. Accordingly, it is not necessary to add the norm of parameters $\hat{W}$ and $\tilde{W}$ as regularizers. 

The first term $L$ is the mapping loss function to measure the distance between logical label and recovered label distribution. According to \cite{xu2018LE}, it is reasonable to select the least squared (LS) function:
\begin{equation}
\begin{aligned} L(\hat{W}) &=\sum_{i=1}^{n}\left\|\hat{W} \phi_{i}-\boldsymbol{l}_{i}\right\|^{2} \\ &=\operatorname{tr}\left[(\hat{W} \Phi-L)^{\top}(\hat{W} \Phi-L)\right] \end{aligned}
\end{equation}
where $\Phi=\left[\phi_{1}, \ldots, \phi_{n}\right]$ and $t r(\cdot)$ is the trace of a matrix defined by the sum of diagonal elements. The second term $R$ is the reconstruction loss function to measure the similarity between the input feature data and the reconstructed one from the output data of LE. Similar to the mapping loss function, the reconstruction loss function is defined as follows:
\begin{equation}
\begin{aligned} 
R(\tilde{W}) &=\sum_{i=1}^{n}\left\|\phi_{i}-\tilde{W} l_{i}\right\|^{2} \\ &=\operatorname{tr}\left[(\Phi-\tilde{W} L)^{T}(\Phi-\tilde{W} L)\right] 
\end{aligned}
\end{equation}

To further simplify the model, it is reasonable to 
consider the tied weights \cite{boureau2008sparseFeature} as follows:
\begin{equation}
\tilde{W}^{*}=\hat{W}^{T}
\end{equation}
where $\tilde{W}^{*}$ is the best reconstruction parameter to be obtained.
Then the Eq. (\ref{eq:obj_le_1}) is rewritten as:
\begin{equation}
\label{eq:obj_le}
\min _{\hat{W}} L(\hat{W})+\alpha R(\hat{W})+\lambda \Omega(\hat{W})
\end{equation}

To obtain desired results, the manifold regularization $\Omega$ is designed to capture the topological consistency between feature space and label space, which can fully exploit the hidden label importance from the input instances. Before presenting this term, it is required to introduce the similarity matrix $A$, whose element is defined as:
\begin{equation}
a_{i j}=\left\{\begin{array}{cc}{\exp \left(-\frac{\left\|x_{i}-x_{j}\right\|^{2}}{2 \sigma^{2}}\right)} & {\text { if } x_{j} \in N(i)} \\ {0} & {\text { otherwise }}\end{array}\right.
\end{equation}
where $N(i)$ denotes the set of $K$-nearest neighbors for the instance $x_{i}$, and $\sigma>0$ is the hyper parameter fixed to be 1 in this paper. Inspired by the smoothness assumption \cite{zhu2005semiGraph}, the more correlated two instances are, the closer are the corresponding recovered label distribution, and vice versa. Accordingly, it is reasonable to design the following manifold regularization:
\begin{equation}
\begin{aligned} \Omega(\hat{W}) &=\sum_{i, j} a_{i j}\left\|d_{i}-d_{j}\right\|^{2}=\operatorname{tr}\left(D G D^{\top}\right) \\ &=\operatorname{tr}\left(\hat{W} \Phi G \Phi^{\top} \hat{W}^{\top}\right) \end{aligned}
\end{equation}
where $d_{i}=\hat{W} \phi_{i}$ indicates the recovered label distribution, and $G=\hat{A}-A$ is the Laplacian matrix. Note that the similarity matrix $A$ is asymmetric so that the element of diagonal matrix $\hat{A}$ element is defined as ${\hat a_{ii}} = \sum\nolimits_{j = 1}^n {{(a_{ij}+a_{ji})/2}}$,

By substituting Eqs. (2), (3) and (7) into Eq. (5), the mapping and reconstruction loss function is defined on parameter $\hat{W}$ as follows:
\begin{equation}
\label{eq:fianl_obj_le}
\begin{aligned}
& T(\hat{W})=\operatorname{tr}\left((\hat{W} \Phi-L)^{T}(\hat{W} \Phi-L)\right)\\
& +\alpha \operatorname{tr}\left(\left(\Phi-\hat{W}^{T} L\right)^{T}\left(\Phi-\hat{W}^{T} L\right)\right)\\
& +\lambda \operatorname{tr}\left(\hat{W} \Phi G \Phi^{T} \hat{W}^{T}\right)
\end{aligned}
\end{equation}
Actually, Eq.(\ref{eq:fianl_obj_le}) can be easily optimized by a well-known method called limited-memory quasi-Newton method (L-BFGS) \cite{yuan1991modifiedBFGS}. This method achieves the optimization by calculating the first-order gradient of $T(\hat{W})$:
\begin{equation}
\begin{aligned}
\frac{\partial T}{\partial \hat{W}}=2 \hat{W} \Phi \Phi^{T}-2 L \Phi^{T}-2 \alpha L \Phi^{T}+2 \alpha L L^{T} \hat{W}\\
+\lambda \hat{W} \Phi G^{T} \Phi^{T}+\lambda \hat{W} \Phi G \Phi^{T}
\end{aligned}
\end{equation}

\subsection{Bi-directional for Label Distribution Learning}

Given dataset $S=\left\{\left(x_{1}, d_{1}\right)\left(x_{2}, d_{2}\right), \cdots,\left(x_{n}, d_{n}\right)\right\}$ whose label is the real-valued format, LDL aims to build a mapping function $f : \mathcal{X} \rightarrow \mathcal{D}$ from the instances to the label distributions, where $x_{i} \in \mathcal{X}$ denotes the $i$-th instance and $d _ { i } = \left\{ d _ { x  _ { i } } ^ { y _ { 1 } } , d _ { x  _ { i } } ^ { y _ { 2 } } , \cdots , d _ { x _ { i } } ^ { y _ { c } } \right\} \in \mathcal{D} $ indicates the $i$-th label distribution of instance. Note that $d_{ x }^{ y }$ accounts for the description degree of $y$ to $x$ rather than the probability that label tags correctly. All the labels can describe each instance completely, so it is reasonable that $d _ { x } ^ { y } \in [ 0,1 ]$ and $\sum _ { y } d _ { x } ^ { y } = 1$. 

As mentioned before, most of LDL methods suffer from the mapping information loss due to the unidirectional projection of loss function. Fortunately, bidirectional projections can extremely preserve the information of input matrix. Accordingly, the goal of our specific BD-LDL algorithm is to determine a mapping parameter $\theta$ and a reconstruction parameter $\tilde{\theta}$ from training set so as to make the predicted label distribution and the true one as similar as possible. Therefore, the new loss function integrates the mapping error with the reconstruction error $R(\tilde{\theta},S)$ as follows:
\begin{equation}
\min _{\theta, \tilde{\theta}} L(\theta, S)+\lambda_{1} R(\tilde{\theta}, S)+\frac{1}{2} \lambda_{2} \Omega(\theta, S)+\frac{1}{2} \lambda_{2} \Omega(\tilde{\theta}, S)
\end{equation}
where ${\theta}$ denotes the mapping parameter, $\tilde{\theta}$ indicates the reconstruction parameter, $\Omega$ is a regularization to control the complexity of the output model to avoid over-fitting, $\lambda_{1}$ and $\lambda_{2}$ are two parameters to balance these four terms.

There are various candidate functions to measure the difference between two distributions such as the Euclidean distance, the Kullback-Leibler (K-L) divergence and the Clark distance etc. Here, we choose the Euclidean distance:
\begin{equation}
L(\theta, S)=\|X \theta-D\|_{F}^{2}
\end{equation}
where $\theta \in R^{d \times c}$ is the mapping parameter to be optimized, and $\|\cdot\|_{F}^{2}$ is the Frobenius norm of a matrix.
For simplification, it is reasonable to consider tied weights \cite{boureau2008sparseFeature} as follows:
\begin{equation}
\tilde{\theta}^{*}=\theta^{T}
\end{equation}
Similarly, the objective function is simplified as follows:
\begin{equation}
\min _{\theta} L(\theta, S)+\lambda_{1} R(\theta, S)+\lambda_{2} \Omega(\theta, S)
\end{equation}
where the term $R(\theta, S)=\left\|X-D \theta^{T}\right\|_{F}^{2}$ denotes the simplified reconstruction error.
As for the second term in objective function, we adopt the F-norm to implement it:
\begin{equation}
\Omega(\theta, S)=\|\theta\|_{F}^{2}
\end{equation}
Substituting Eqs. (11) and (14) into Eq. (13) yields the objective function:
\begin{equation}
\min _{\theta}\|X \theta-D\|_{F}^{2}+\lambda_{1}\left\|X-D \theta^{T}\right\|_{F}^{2}+\lambda_{2}\|\theta\|_{F}^{2}
\end{equation}

Before optimization, the trace properties $\operatorname{tr}(X)=\operatorname{tr}\left(X^{T}\right)$ and $\operatorname{tr}\left(D \theta^{T}\right)=\operatorname{tr}\left(\theta D^{T}\right)$ are applied for the re-organization of objective function:
\begin{equation}
\min _{\theta}\|X \theta-D\|_{F}^{2}+\lambda_{1}\left\|X^{T}-\theta D^{T}\right\|_{F}^{2}+\lambda_{2}\|\theta\|_{F}^{2}.
\label{eq:der0}
\end{equation}
Then, for optimization, we can simply take a derivative of Eq. (\ref{eq:der0}) with respective to the parameter $\theta$ and set it zero:
\begin{equation}
{X^{T}(X \theta-D)-\lambda_{1}\left(X^{T}-\theta D^{T}\right) D+\lambda_{2} \theta=0} 
\label{eq:der}
\end{equation}
Obviously, Eq. (\ref{eq:der}) can be transformed into the following equivalent formulation:
\begin{equation}
{\left(X^{T} X+\lambda_{2} I\right) \theta+\lambda_{1} \theta D^{T} D=X^{T} D+\lambda_{1} X^{T} D}
\label{eq:bd_ldl_final}
\end{equation}
Denote $A=X^{T} X+\lambda_{2} I$, $B=\lambda_{1} D^{T} D$ and $C=\left(1+\lambda_{1}\right) X^{T} D$,
Eq. (\ref{eq:bd_ldl_final}) can be rewritten as the following formulation:
\begin{equation}
\label{eq:AB=C}
A \theta+\theta B=C
\end{equation}
\begin{algorithm}[htb]  
	\caption{BD-LDL Algorithm}  
	\label{alg:BD-LDL}  
	\begin{algorithmic}[1]  
		\Require  
		$X$: $n \times d$ training feature matrix;
		
		$D$: $n \times c$ labeled distribution matrix; 
		
		\Ensure  
		$\theta$: $d \times c$ projection parameter
		
		\State Initial $\theta^{0}$, $\lambda_{1}$, $\lambda_{2}$ and $t=0$; 
		\Repeat
		\State Compute $A$, $B$ and $C$ in Eq.(\ref{eq:AB=C})
		
		\State Perform Cholesky factorization to gain $P$ and $Q$
		
		\State Perform SVD on $P$ and $Q$
		
		\State Update $\tilde{\theta}^{t+1}$ via Eqs.(\ref{eq:step3}) and (\ref{eq:element})
		
		\State Update $\theta^{t+1}$ via Eqs.(\ref{eq:the_best_theta})
		
		\Until{Stopping criterion is satisfied}  
		
	\end{algorithmic}  
\end{algorithm}

Although Eq. (\ref{eq:AB=C}) is the well-known Sylvester equation which can be solved by existing algorithm in MATLAB, the computational cost corresponding solution is not ideal. Thus, following \cite{zhu2017a}, we effectively solve Eq. (\ref{eq:AB=C}) with Cholesky factorization \cite{golub1996matrix} as well the Singular Value Decomposition (SVD). Firstly, two positive semi-definite matrix $A$ and $B$ can be factorized as:
\begin{equation}
\label{eq:Cholesky}
\begin{aligned}
&A=P^{T} \times P\\
&B=Q \times Q^{T}
\end{aligned}
\end{equation}
where $P$ and $Q$ are the triangular matrix which can be further decomposed via SVD as: 
\begin{equation}
\label{eq:SVD}
\begin{aligned}
& P=U_{1} \Sigma_{1} V_{1}^{T} \\
& Q=U_{2} \Sigma_{2} V_{2}^{T}
\end{aligned}
\end{equation}
Substituting Eqs. (\ref{eq:Cholesky}) and (\ref{eq:SVD}) into Eq. (\ref{eq:AB=C}) yields:

\begin{equation}
\label{eq:step1}
V_{1} \Sigma_{1}^{T} U_{1}^{T} U_{1} \Sigma_{1} V_{1}^{T} \theta +\theta U_{2} \Sigma_{2} V_{2}^{T} V_{2} \Sigma_{2}^{T} U_{2}^{T}=C
\end{equation}
Since $U_{1}$, $U_{2}$, $V_{1}$ and $V_{2}$ are the unitray matrix, Eq. (\ref{eq:step1}) can be rewritten as :
\begin{equation}
\label{eq:step2}
V_{1} \Sigma_{1}^{T} \Sigma_{1} V_{1}^{T} \theta +\theta U_{2} \Sigma_{2} \Sigma_{2}^{T} U_{2}^{T}=C
\end{equation}
We multiplying $V_{1}^{T}$ and $U_{2}$ to both sides of Eq. (\ref{eq:step2}) to obtain the following equation:
\begin{equation}
\label{eq:step3}
\tilde{\Sigma}_{1} \tilde{\theta} + \tilde{\theta} \tilde{\Sigma}_{2} = E
\end{equation}
where $\tilde{\Sigma}_{1}=\Sigma_{1}^{T} \Sigma_{1}$, $\tilde{\Sigma}_{2}=\Sigma_{2} \Sigma_{2}^{T}$, $E=V_{1}^{T} C U_{2}$ and $\tilde{\theta}=V_{1}^{T} \theta U_{2}$.

For both $\tilde{\Sigma}_{1}$ and $\tilde{\Sigma}_{2}$ are the diagonal matrix, we can directly attain $\tilde{\theta}$ whose element is defined as:
\begin{equation}
\label{eq:element}
\tilde{\theta}_{i j}=\frac{e_{i j}}{\tilde{\sigma}_{i i}^{1}+\tilde{\sigma}_{j j}^{2}}
\end{equation}
where $\tilde{\sigma}_{i i}^{1}$ and $\tilde{\sigma}_{i i}^{2}$ can be calculated by eigenvalues of $P$ and $Q$ respectively, and $e_{i j}$ is the \textit{i,j}-th elment of matrix $E$. Accordingly, $\theta$ can be obtained by:

\begin{equation}
\label{eq:the_best_theta}
\theta=V_{1} \tilde{\theta} U_{2}^{T}
\end{equation}

We briefly summarize the procedure of the proposed BD-LDL in Algorithm 1.
\begin{table}[]
	\centering
	\caption{Statistics of 13 datasets used in comparison experiment}\smallskip
	\resizebox{0.55\columnwidth}{!}{
		\begin{tabular}{lllll}
			\toprule
			Index & Data\ Set & \# Examples & \# Features & \# Labels \\
			\midrule
			1 & Yeast-alpha & 2,465 & 24 & 18 \\
			2 & Yeast-cdc & 2,465 & 24 & 15 \\
			3 & Yeast-cold & 2,465 & 24 & 4 \\
			4 & Yeast-diau & 2,465 & 24 & 7 \\
			5 & Yeast-dtt & 2,465 & 24 & 4 \\
			6 & Yeast-elu & 2,465 & 24 & 14 \\
			7 & Yeast-heat & 2,465 & 24 & 6 \\
			8 & Yeast-spo & 2,465 & 24 & 6 \\
			9 & Yeast-spo5 & 2,465 & 24 & 3 \\
			10 & Yeast-spoem & 2,465 & 24 & 2 \\
			11 & Natural\ Scene & 2,000 & 294 & 9 \\
			12 & Movie & 7,755 & 1,869 & 5 \\
			13 & SBU\_3DFE & 2,500 & 243 & 6 \\
			\bottomrule
		\end{tabular}
	}
	\label{tab:Data}
\end{table}

\section{Experiments}
\subsection{Datasets and Measurement}
We conducted extensive experiments on 13 real-world datasets collected from biological experiments \cite{eisen1998Genecluster}, facial expression images \cite{lyons1998coding-facial}, natural scene images, and movies.
The output of both LE and LDL are in the format of label distribution vectors. In contrast to the results of SLL and MLL, the label distribution vectors should be evaluated with diverse measurements. We naturally select six criteria that are most commonly used, i.e., Chebyshev distance (Chebeyshev), Clark distance (Clark), Canberra metric (Canberra), Kullback–Leibler divergence (K-L), Cosine coefficient (Cosine), and Intersection similarity (Intersec). The first four functions are always used to measure distance between groud-truth label distribution $D$ and the predicted one $\widehat{D}$, whereas the last two are similarity measurements.The specifications of criteria and used data sets can be found in Tables \ref{tab:Evaluation measurements} and \ref{tab:Data}.
\begin{table}[]
	\large
	\centering
	\caption{Evaluation Measurements}\smallskip
	\resizebox{0.55\columnwidth}{!}{
		\begin{tabular}{l|l|l}
			\toprule
			\  & Name & Defination  \\
			\midrule
			Distance & Chebyshev $\downarrow$ & $D i s_{1}(D, \widehat{D})=\max _{i}\left|d_{i}-\widehat{d}_{i}\right|$ \\
			\ & Clark $\downarrow$ & $D i s_{2}(D, \widehat{D})=\sqrt{\sum_{i=1}^{c} \frac{\left(d_{i}-\widehat{d_{i}}\right)^{2}}{\left(d_{i}+\widehat{d_{i}}\right)^{2}}}$ \\
			\ & Canberra $\downarrow$ & $D i s_{3}(D, \widehat{D})=\sum_{i=1}^{c} \frac{\left|d_{i}-\widehat{d}_{i}\right|}{d_{i}+\widehat{d}_{i}}$ \\
			\midrule
			Similarity & Intersaction $\uparrow$ & $S i m _{1}(D, \widehat{D})=\sum_{i=1}^{c} \min \left(d_{i}, \widehat{d}_{i}\right)$ \\
			\ & Cosine $\uparrow$ & $S i m_{2}(D, \widehat{D})=\frac{\sum_{i=1}^{c} d_{i} \widehat{d}_{i}}{\sqrt{\left(\sum_{i=1}^{c} d_{i}^{2}\right)\left(\sum_{i=1}^{c} \widehat{d}_{i}^{2}\right)}}$ \\
			\toprule
		\end{tabular}
	}
	\label{tab:Evaluation measurements}
\end{table}

\begin{table}[]
	\label{tab:LE-CHEB}
	\centering
	\caption{Comparison Performance(rank) of Different LE Algorithms Measured by Chebyshev $\downarrow$}\smallskip
	\resizebox{0.8\columnwidth}{!}{
		\begin{tabular}{lllllll}
		    \toprule
			Datasets & Ours & FCM & KM & LP & ML & GLLE\\[2pt]
			\midrule
			Yeast-alpha & \bf 0.0208(1) & 0.0426(4) & 0.0588(6) & 0.0401(3) & 0.0553(5) & 0.0310(2)\\
			Yeast-cdc & \bf 0.0231(1) & 0.0513(4) & 0.0729(6) & 0.0421(3) & 0.0673(5) & 0.0325(2)\\
			Yeast-cold & \bf 0.0690(1) & 0.1325(4) & 0.2522(6) & 0.1129(3) & 0.2480(5) & 0.0903(2)\\
			Yeast-diau & \bf 0.0580(1) & 0.1248(4) & 0.2500(6) & 0.0904(3) & 0.1330(5) & 0.0789(2)\\
			Yeast-dtt & \bf 0.0592(1) & 0.0932(3) & 0.2568(5) & 0.1184(4) & 0.2731(6) & 0.0651(2)\\
			Yeast-elu & \bf 0.0256(1) & 0.0512(4) & 0.0788(6) & 0.0441(3) & 0.0701(5) & 0.0287(2)\\
			Yeast-heat & \bf 0.0532(1) & 0.1603(4) & 0.1742(5) & 0.0803(3) & 0.1776(6) & 0.0563(2)\\
			Yeast-spo & \bf 0.0641(1) & 0.1300(4) & 0.1753(6) & 0.0834(3) & 0.1722(5) & 0.0670(2)\\
			Yeast-spo5 & 0.1017(2) & 0.1622(4) & 0.2773(6) & 0.1142(3) & 0.2730(5) & \bf 0.0980(1)\\
			Yeast-spoem & \bf 0.0921(1) & 0.2333(4) & 0.4006(6) & 0.1632(3) & 0.3974(5) & 0.1071(2)\\
			Natural\_Scene & 0.3355(5) & 0.3681(6) & 0.3060(3) & \bf 0.2753(1) & 0.2952(2) & 0.3349(4)\\
			Movie & \bf 0.1254(1) & 0.2302(4) & 0.2340(6) & 0.1617(3) & 0.2335(5) & 0.1601(2)\\
			SUB\_3DFE & \bf 0.1285(1) & 0.1356(3) & 0.2348(6) & 0.1293(2) & 0.2331(5) & 0.1412(4)\\
			\midrule
			Avg. Rank & 1.38 & 4.00 & 5.62 & 2.84 & 4.92 & 2.23 \\
			\bottomrule
		\end{tabular}
	}\label{BD_LDL_RESULTS_1}
\end{table}
\begin{table}[]
	\label{tab:LE-COSINE}
	\centering
	\caption{Comparison Performance(rank) of Different LE Algorithms Measured by Cosine $\uparrow$}\smallskip
	\resizebox{0.8\columnwidth}{!}{
		\begin{tabular}{lllllll}
		    \toprule
			Datasets & Ours & FCM & KM & LP & ML & GLLE\\[2pt]
		    \midrule
			Yeast-alpha & \bf 0.9852(1) & 0.9221(3) & 0.8115(5) & 0.9220(4) & 0.7519(6) & 0.9731(2)\\
			Yeast-cdc & \bf 0.9857(1) & 0.9236(3) & 0.7541(6) & 0.9162(4) & 0.7591(5) & 0.9597(2)\\
			Yeast-cold & \bf 0.9804(1) & 0.9220(4) & 0.7789(6) & 0.9251(3) & 0.7836(5) & 0.9690(2)\\
			Yeast-diau & \bf 0.9710(1) & 0.8901(4) & 0.7990(6) & 0.9153(3) & 0.8032(5) & 0.9397(2)\\
			Yeast-dtt & \bf 0.9847(1) & 0.9599(3) & 0.7602(6) & 0.9210(4) & 0.7631(5) & 0.9832(2)\\
			Yeast-elu & \bf 0.9841(1) & 0.9502(3) & 0.7588(5) & 0.9110(4) & 0.7562(6) & 0.9813(2)\\
			Yeast-heat & \bf 0.9803(1) & 0.8831(4) & 0.7805(6) & 0.9320(3) & 0.7845(5) & 0.9800(2)\\
			Yeas-spo & \bf 0.9719(1) & 0.9092(4) & 0.8001(6) & 0.9390(3) & 0.8033(5) & 0.9681(2)\\
			Yeast-spo5 & 0.9697(2) & 0.9216(4) & 0.8820(6) & 0.9694(3) & 0.8841(5) & \bf 0.9713(1)\\
			Yeast-spoem & \bf 0.9761(1) & 0.8789(4) & 0.8122(6) & 0.9500(3) & 0.8149(5) & 0.9681(2)\\
			Natural\_Scene & 0.7797(4) & 0.5966(6) & 0.7488(5) & 0.8602(2) & \bf 0.8231(1) & 0.7822(3)\\
			Movie & \bf 0.9321(1) & 0.7732(6) & 0.8902(4) & 0.9215(2) & 0.8153(5) & 0.9000(3)\\
			SBU\_3DFE & \bf 0.9233(1) & 0.9117(3) & 0.8126(6) & 0.9203(2) & 0.8150(5) & 0.9000(4)\\
			\midrule
			Avg. Rank & 1.31 & 3.92 & 5.62 & 3.08 & 4.85 & 2.23 \\
			\bottomrule
		\end{tabular}
	}\label{BD_LDL_RESULTS_2}
\end{table}

\begin{table}
	\label{tab:LDL-CHEB}
	\centering
	\caption{Comparison Results(mean$\pm$std.(rank)) of Different LDL Algorithms Measured by Clark $\downarrow$}\smallskip
	\resizebox{1\textwidth}{!}{
		\begin{tabular}{lllllllll}
		    \toprule
			Datasets & Ours & PT-Bayes & AA-BP & SA-IIS & SA-BFGS & LDL-SCL & EDL-LRL & LDLLC\\ 
			\midrule
			Yeast-alpha	& \bf 0.2097$\pm$0.003(1)	& 1.1541$\pm$0.034(8)	& 0.7236$\pm$0.060(7)	& 0.3053$\pm$0.006(6)	& 0.2689$\pm$0.008(5)	& 0.2098$\pm$0.002(2)	& 0.2126$\pm$0.000(4)	& 0.2098$\pm$0.006(3) \\
			Yeast-cdc	& \bf 0.2017$\pm$0.004(1)	& 1.0601$\pm$0.066(8)	& 0.5728$\pm$0.030(7)	& 0.2932$\pm$0.004(6)	& 0.2477$\pm$0.007(5)	& 0.2137$\pm$0.004(3)	& 0.2046$\pm$2.080(2)	& 0.2163$\pm$0.004(4) \\
			Yeast-cold	& \bf 0.1355$\pm$0.004(1)	& 0.5149$\pm$0.024(8)	& 0.1552$\pm$0.005(7)	& 0.1643$\pm$0.004(8)	& 0.1471$\pm$0.004(5)	& 0.1388$\pm$0.003(2)	& 0.1442$\pm$2.100(4)	& 0.1415$\pm$0.004(3) \\
			Yeast-diau	& \bf 0.1960$\pm$0.006(1)	& 0.7487$\pm$0.042(8)	& 0.2677$\pm$0.010(7)	& 0.2409$\pm$0.006(6)	& 0.2201$\pm$0.002(5)	& 0.1986$\pm$0.002(2)	& 0.2011$\pm$0.003(4)	& 0.2010$\pm$0.006(3) \\
			Yeast-dtt	& 0.0964$\pm$0.004(2)	& 0.4807$\pm$0.040(8)	& 0.1206$\pm$0.008(7)	& 0.1332$\pm$0.003(8)	& 0.1084$\pm$0.003(6)	& 0.0989$\pm$0.001(4)	& 0.0980$\pm$1.600(3)	& \bf 0.0962$\pm$0.006(1) \\
			Yeast-elu	& \bf 0.1964$\pm$0.004(1)	& 1.0050$\pm$0.041(8)	& 0.5246$\pm$0.028(7)	& 0.2751$\pm$0.006(6)	& 0.2438$\pm$0.008(5)	& 0.2015$\pm$0.002(3)	& 0.2029$\pm$0.023(4)	& 0.1994$\pm$0.006(2) \\
			Yeast-heat	& \bf 0.1788$\pm$0.005(1)	& 0.6829$\pm$0.026(8)	& 0.2261$\pm$0.010(7)	& 0.2260$\pm$0.005(6)	& 0.1998$\pm$0.003(5)	& 0.1826$\pm$0.003(2)	& 0.1826$\pm$0.003(2)	& 0.1854$\pm$0.004(4) \\
			Yeast-spo	& \bf 0.2456$\pm$0.008(1)	& 0.6686$\pm$0.040(8)	& 0.2950$\pm$0.010(7)	& 0.2759$\pm$0.006(6)	& 0.2639$\pm$0.003(5)	& 0.2503$\pm$0.002(4)	& 0.2480$\pm$0.685(2)	& 0.2500$\pm$0.008(3) \\
			Yeast-spo5	& \bf 0.1785$\pm$0.007(1)	& 0.4220$\pm$0.020(8)	& 0.1870$\pm$0.005(3)	& 0.1944$\pm$0.009(6)	& 0.1962$\pm$0.001(7)	& 0.1881$\pm$0.004(4)	& 0.1915$\pm$0.020(5)	& 0.1837$\pm$0.007(2) \\
			Yeast-spoem	& \bf 0.1232$\pm$0.005(1)	& 0.3065$\pm$0.030(8)	& 0.1890$\pm$0.012(7)	& 0.1367$\pm$0.007(6)	& 0.1312$\pm$0.001(3)	& 0.1316$\pm$0.005(4)	& 0.1273$\pm$0.054(2)	& 0.1320$\pm$0.008(5) \\
			Natural\_Scene	& \bf 2.3612$\pm$0.541(1)	& 2.5259$\pm$0.015(8)	& 2.4534$\pm$0.018(4)	& 2.4703$\pm$0.019(6)	& 2.4754$\pm$0.013(7)	& 2.4580$\pm$0.012(5)	& 2.4519$\pm$0.005(2)	& 2.4456$\pm$0.019(3) \\
			Movie	& \bf 0.5211$\pm$0.606(1)	& 0.8044$\pm$0.010(8)	& 0.6533$\pm$0.010(6)	& 0.5783$\pm$0.007(5)	& 0.5750$\pm$0.011(4)	& 0.5543$\pm$0.007(3)	& 0.6956$\pm$0.041(7)	& 0.5289$\pm$0.008(2) \\
			SBU\_3DFE	& 0.3540$\pm$0.010(2)	& 0.4137$\pm$0.010(5)	& 0.4454$\pm$0.020(8)	& 0.4156$\pm$0.012(7)	& \bf 0.3465$\pm$0.006(1)	& 0.3546$\pm$0.002(3)	& 0.3556$\pm$0.006(4)	& 0.4145$\pm$0.006(6) \\
			\midrule
			Avg. Rank & 1.15 & 7.77 & 6.46 & 6.31 & 4.85 & 3.15 & 3.46 & 3.15 \\
			\bottomrule
		\end{tabular}
	}\label{BD_LDL_RESULTS_1}
\end{table}
\begin{table}
	\label{tab:LDL-COSINE}
	\centering
	\caption{Comparison Results(mean$\pm$std.(rank)) of Different LDL Algorithms Measured by Cosine $\uparrow$}\smallskip
	\resizebox{1\textwidth}{!}{
		\begin{tabular}{lllllllll}
		    \toprule
			Datasets & Ours & PT-Bayes & AA-BP & SA-IIS & SA-BFGS & LDL-SCL & EDL-LRL & LDLLC\\ 
			\midrule
			Yeast-alpha & \bf 0.9947$\pm$0.000(1) & 0.8527$\pm$0.005(8) & 0.9482$\pm$0.007(7) & 0.9879$\pm$0.000(6) & 0.9914$\pm$0.000(5) & 0.9945$\pm$0.000(3) & 0.9945$\pm$0.000(4) & 0.9946$\pm$0.000(2)\\
			Yeast-cdc & \bf 0.9955$\pm$0.000(1) & 0.8544$\pm$0.012(8) & 0.9590$\pm$0.003(7) & 0.9871$\pm$0.000(6) & 0.9913$\pm$0.000(4) & 0.9904$\pm$0.000(5) & 0.9939$\pm$8.070(2) & 0.9932$\pm$0.000(3)\\
			Yeast-cold & \bf 0.9893$\pm$0.001(1) & 0.8884$\pm$0.008(8) & 0.9859$\pm$0.001(6) & 0.9838$\pm$0.000(7) & 0.9871$\pm$0.000(5) & 0.9886$\pm$0.000(3) & 0.9892$\pm$0.034(2) & 0.9883$\pm$0.001(4)\\
			Yeast-diau & \bf 0.9884$\pm$0.001(1) & 0.8644$\pm$0.007(8) & 0.9860$\pm$0.000(5) & 0.9821$\pm$0.000(7) & 0.9853$\pm$0.000(6) & 0.9880$\pm$0.000(2) & 0.9876$\pm$0.063(4) & 0.9878$\pm$0.001(3)\\
			Yeast-dtt & \bf 0.9943$\pm$0.000(1) & 0.8976$\pm$0.012(8) & 0.9909$\pm$0.001(6) & 0.9889$\pm$0.000(7) & 0.9928$\pm$0.000(5) & 0.9939$\pm$0.000(3) & 0.9940$\pm$0.021(2) & 0.9939$\pm$0.001(4)\\
			Yeast-elu & \bf 0.9942$\pm$0.000(1) & 0.8600$\pm$0.008(8) & 0.9623$\pm$0.003(7) & 0.9876$\pm$0.000(6) & 0.9912$\pm$0.000(5) & 0.9939$\pm$0.000(3) & 0.9938$\pm$0.001(4) & 0.9940$\pm$0.000(2)\\
			Yeast-heat & \bf 0.9884$\pm$0.001(1) & 0.8655$\pm$0.008(8) & 0.9814$\pm$0.001(6) & 0.9810$\pm$0.000(7) & 0.9857$\pm$0.000(5) & 0.9880$\pm$0.000(2) & 0.9880$\pm$0.029(3) & 0.9876$\pm$0.001(4)\\
			Yeast-spo & \bf 0.9776$\pm$0.001(1) & 0.8672$\pm$0.010(8) & 0.9686$\pm$0.003(7) & 0.9718$\pm$0.001(6) & 0.9745$\pm$0.000(5) & 0.9768$\pm$0.000(4) & 0.9772$\pm$0.010(2) & 0.9770$\pm$0.001(3)\\
			Yeast-spo5 & \bf 0.9753$\pm$0.002(1) & 0.8968$\pm$0.010(8) & 0.9731$\pm$0.001(4) & 0.9706$\pm$0.002(7) & 0.9710$\pm$0.000(6) & 0.9732$\pm$0.001(3) & 0.9723$\pm$0.007(5) & 0.9743$\pm$0.002(2)\\
			Yeast-spoem & \bf 0.9803$\pm$0.001(1) & 0.9187$\pm$0.010(8) & 0.9728$\pm$0.003(7) & 0.9764$\pm$0.001(6) & 0.9786$\pm$0.000(3) & 0.9784$\pm$0.001(4) & 0.9796$\pm$0.008(2) & 0.9784$\pm$0.002(5)\\
			Natural\_Scene & \bf 0.7637$\pm$0.015(1) & 0.5583$\pm$0.006(8) & 0.6954$\pm$0.014(7) & 0.6986$\pm$0.008(6) & 0.7144$\pm$0.008(5) & 0.7442$\pm$0.007(3) & 0.7624$\pm$0.003(2) & 0.7486$\pm$0.014(4)\\
			Movie & \bf 0.9385$\pm$0.002(1) & 0.8495$\pm$0.003(8) & 0.8767$\pm$0.006(7) & 0.9089$\pm$0.002(4) & 0.8780$\pm$0.004(5) & 0.9205$\pm$0.002(3) & 0.8780$\pm$0.005(6) & 0.9381$\pm$0.003(2)\\
			SUB\_3DFE & \bf 0.9644$\pm$0.004(1) & 0.9167$\pm$0.004(8) & 0.9181$\pm$0.005(7) & 0.9202$\pm$0.004(5) & 0.9482$\pm$0.001(3) & 0.9436$\pm$0.000(4) & 0.9636$\pm$0.002(2) & 0.9198$\pm$0.002(6)\\
			\midrule
			Avg. Rank & 1.00 & 8.00 & 6.38 & 6.15 & 4.77 & 3.54 & 3.08 & 3.38 \\
			\bottomrule
		\end{tabular}
	}\label{BD_LDL_RESULTS_2}
\end{table}
\subsection{Methodology}
To show the effectiveness of the proposed methods, we conducted comprehensive experiments on the aforementioned datasets. For LE, the proposed BD-LE method is compared with five classical LE approaches presented in \cite{xu2018LE}, i.e., FCM, KM, LP, ML, and GLLE. The hyper-parameter in the FCM method is set to 2. We select the Gaussian Kernel as the kernel function in the KM algorithm. For GLLE, the parameter $\lambda$ is set to 0.01. Moreover, the number of neighbors $K$ is set to $c+1$ in both GLLE and ML.

For the LDL paradigm, the proposed BD-LDL method is compared with eight existing algorithms including PT-Bayes \cite{geng2013LDL}, PT-SVM \cite{geng2014facialadaptive}, AA-KNN \cite{geng2010facialestimation}, AA-BP \cite{geng2013facialestimation}, SA-IIS \cite{geng2013facialestimation}, SA-BFGS \cite{geng2016LDL}, LDL-SCL \cite{zheng2018labelCorrelationSample}, and EDL-LRL \cite{jia2019facialEDLLRL}, to demonstrate its superiority. The first two algorithms are implemented by the strategy of problem transformation. The next two ones are carried out by means of the adaptive method. Finally, from the fifth algorithm to the last one, they are specialized algorithms. In particular, the LDL-SCL and EDL-LRL constitute state-of-art methods recently proposed. We utilized the ``C-SVC'' type in LIBSVM to implement PT-SVM using the RBF kernel with parameters $C=10^{-1}$ and $\gamma=10^{-2}$. We set the hyper-parameter $k$ in AA-kNN to 5. The number of hidden-layer neurons for AA-BP was set to 60. The parameters $\lambda_{1}$, $\lambda_{2}$ and $\lambda_{3}$ in LDL-SCL were all set to $10^{-3}$. Regarding the EDL-LRL algorithm, we set the regularization parameters $\lambda_{1}$ and $\lambda_{2}$ to $10^{-3}$ and $10^{-2}$, respectively. For the intermediate algorithm K-means, the number of cluster was set to 5 according to Jia`s suggestion \cite{jia2018labelCorrelation}. For the BFGS optimization used in SA-BFGS and BD-LDL, parameters $c_{1}$ and $c_{2}$ were set to $10^{-4}$ and 0.9, respectively. 
Regarding the two bi-directional algorigthms, parameters are tuned from the range $10^{\{-4,-3,-2,-1,0,1,2,3\}}$ using grid-search method. The two  parameters in BD-LE $\alpha$ and $\lambda$ are both set to $10^{-3}$. As for BD-LDL, the parameters $\lambda_{1}$ and $\lambda_{2} $ are set to $10^{-3}$ and $10^{-2}$, respectively. Finally, we train the LDL model with the recovered label distributions for further evaluation of BD-LE. The details of parameter selections are shown in the parameter analysis section. And the experiments for the LDL algorithm on every datasets are conducted on a ten-fold cross-validation.

\subsection{Results}
\subsubsection{BD-LE performance}
Tables 1 and 2 present the results of six LE methods on all the datasets. Constrained by the page limit, we have only shown two representative results measured on Chebyshev and Cosine in this paper. 

For each dataset, the results made by a specific algorithm are listed as a column in the tables in accordance with the used matrix. Note that there is always an entry highlighted in boldface. This entry indicates that the algorithm evaluated by the corresponding measurement achieves the best performance. The experimental results are presented in the form of ``score (rank)''; ``score'' denotes the difference between a predicted distribution and the real one measured by the corresponding matrix; ``rank'' is a direct value to evaluate the effectiveness of these compared algorithms. Moreover, the symbol ``$\downarrow$'' means ``the smaller the better'', whereas ``$\uparrow$'' indicates ``the larger the better.''
 \begin{figure}
	\centering
	\includegraphics[width=1\columnwidth]{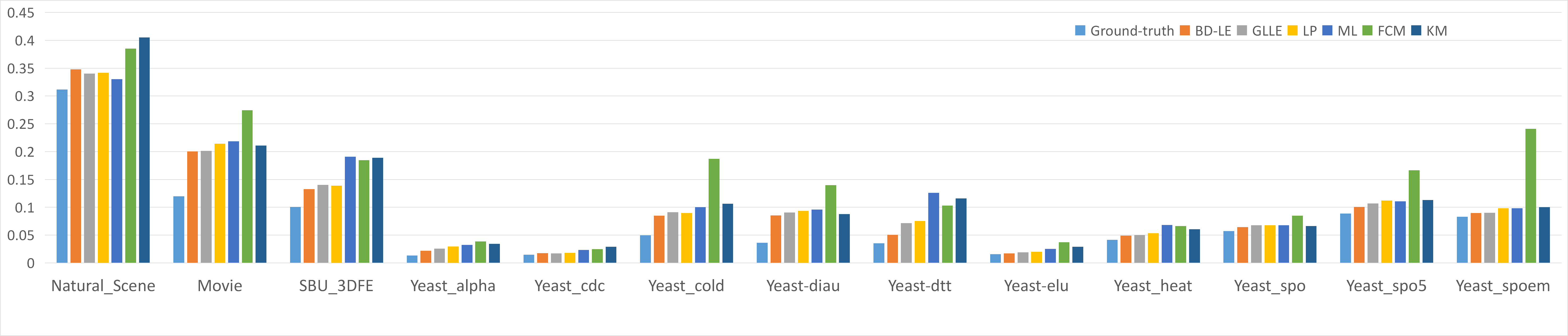}
	\caption{Comparison results of different LE algorithms against the ‘Ground-truth’ used in BD-LDL measured by Chebyshev $\downarrow$}
	\label{fig:BD-LDL-CHEB}
\end{figure}

It is worth noting that given that the LE method is regarded as a pre-processing method, there is no need to run it several times and record the mean as well as the standard deviation. After analyzing the results we obtained, the proposed BD-LE clearly outperforms other LE algorithms in most of the cases and renders sub-optimum performance only in about 4.7\% of cases according to the statistics. In addition, BD-LE achieves better prediction results than GLLE in most of the cases, especially on dataset movie. From Table 1 we can see that the largest dimensional gap between input space and the output one is exactly in dataset moive. This indicates that the reconstruction projection can be added in LE algorithm reasonably. Two specialized algorithms, namely BE-LE and GLLE, rank first in 91.1\% of cases. By contrast, the label distributions are hardly recovered from other four algorithms. This indicates the superiority of utilizing direct similarity or distance as the loss function in LDL and LE problems. In summary, the performance of the five LE algorithms is ranked from best to worst as follows: BD-LE, GLLE, LP, FCM, ML and KM. This proves the effectiveness of our proposed bi-directional loss function for the LE method.
\begin{figure}
	\centering
	\includegraphics[width=1\columnwidth]{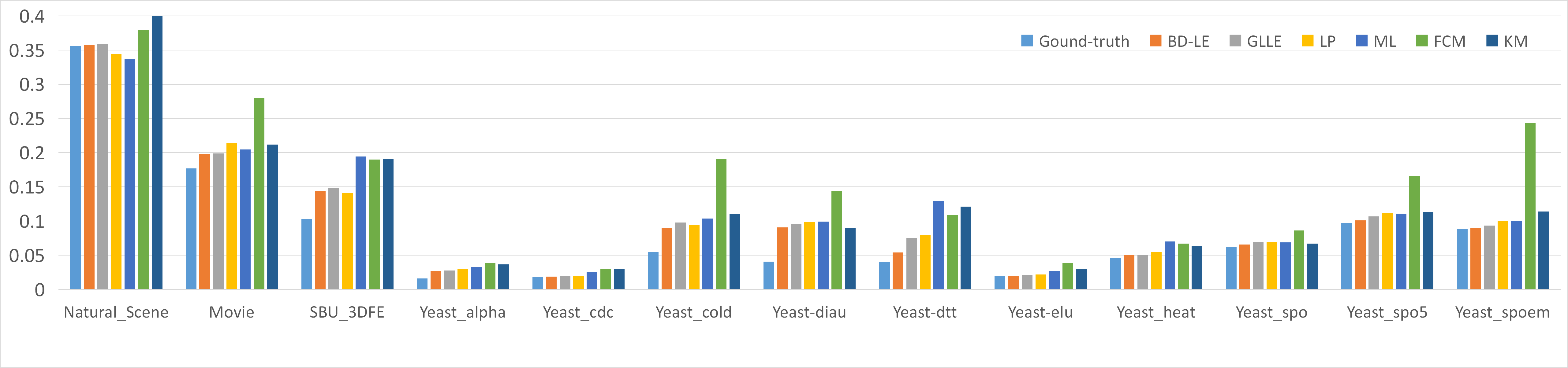}
	\caption{Comparison results of different LE algorithms against the ‘Ground-truth’ used in SA-BFGS measured by Chebyshev $\downarrow$}
	\label{fig:SA-BFGS-CHEB}
\end{figure}
\subsubsection{BD-LDL performance}

As for the performance of BD-LDL method, we show the numerical result on 13 real-world datasets over the measurement Clark and Cosine in Tables 3 and 4 with the format ``mean$\pm$std (rank)'' similarly, and the item in bold in every row represents the best performance. One may observe that our algorithm BD-LDL outperforms other classical LDL algorithms in most cases. When measured by Cosine, it is vividly shown that BD-LDL achieves the best performance on every datasets, which strongly demonstrates the effectiveness of our proposed method. Besides, it can be found from Table 4 that although LDLLC and SA-BFGS obtains the best result on dataset Yeast-dtt and SBU\_3DFE respectively when measured with Chebyshev, BD-LDL still ranks the second place. It also can be seen from the results that two PT and AP algorithms perform poorly on most cases. This verifies the superiority of utilizing the direct similarity or distance between the predicted label distribution and the true one as the loss function. Moreover, it can be easily seen from the results that our proposed method gains the superior performance over other existing specialized algorithms which ignore considering the reconstruction error. This indicates that such a bi-directional loss function can truly boost the performace of LDL algorithm.
\begin{table}[]
	\label{tab:ablation}
	\centering
	\caption{ Ablation experiments results of UD-LDL and BD-LDL Algorithms Measured by Canberra $\downarrow$ and Intersection $\uparrow$ }\smallskip
	\resizebox{0.8\columnwidth}{!}{
		\begin{tabular}{l|ll|ll} 
		    \toprule
			\multirow{2}{*}{Dataset}  & \multicolumn{2}{l|}{$\qquad$$\qquad$Canberra $\downarrow$}   & \multicolumn{2}{l}{$\qquad$$\qquad$Intersection $\uparrow$} \\
			\cline{2-5}
			&  UD-LDL & BD-LDL & UD-LDL & BD-LDL \\
			\midrule
			\centering
			Yeast-alpha	& 0.7980$\pm$0.013	& \bf 0.6013$\pm$0.011	& 0.8915 $\pm$ 0.001	& \bf 0.9624$\pm$0.001 \\
			Yeast-cdc	& 0.9542$\pm$0.012	& \bf 0.6078$\pm$0.012	& 0.8869 $\pm$ 0.001	& \bf 0.9580$\pm$0.001 \\
			Yeast-cold	& 0.4515$\pm$0.010	& \bf 0.2103$\pm$0.007	& 0.8779 $\pm$ 0.002	& \bf 0.9430$\pm$0.002 \\
			Yeast-diau	& 0.6751$\pm$0.010	& \bf 0.4220$\pm$0.013	& 0.8338 $\pm$ 0.001	& \bf 0.9414$\pm$0.002 \\
			Yeast-dtt	& 0.3724$\pm$0.010	& \bf 0.1659$\pm$0.006	& 0.8775 $\pm$ 0.002	& \bf 0.9590$\pm$0.001 \\
			Yeast-elu	& 0.8005$\pm$0.007	& \bf 0.5789$\pm$0.011	& 0.8776 $\pm$ 0.001	& \bf 0.9591$\pm$0.001 \\
			Yeast-heat	& 0.5706$\pm$0.010	& \bf 0.3577$\pm$0.010	& 0.8591 $\pm$ 0.002	& \bf 0.9412$\pm$0.002 \\
			Yeast-spo	& 0.7075$\pm$0.016	& \bf 0.5036$\pm$0.019	& 0.8301 $\pm$ 0.003	& \bf 0.9171$\pm$0.003 \\
			Yeast-spo5	& 0.4834$\pm$0.010	& \bf 0.2745$\pm$0.011	& 0.8184 $\pm$ 0.003	& \bf 0.9112$\pm$0.003 \\
			Yeast-spoem	& 0.2998$\pm$0.010	& \bf 0.1716$\pm$0.007	& 0.8129 $\pm$ 0.005	& \bf 0.9169$\pm$0.003 \\
			Natural\_Scene	& 6.9653$\pm$0.095	& \bf 0.7319$\pm$0.040	& 0.3822 $\pm$ 0.010	& \bf 0.5395$\pm$0.011 \\
			Movie	& 1.4259$\pm$0.024	& \bf 0.0218$\pm$0.001	& 0.7429 $\pm$ 0.004	& \bf 0.8298$\pm$0.002 \\
			SBU\_3DFE	& 0.8562$\pm$0.021	& \bf 0.0119$\pm$0.001	& 0.8070 $\pm$ 0.004	& \bf 0.8590$\pm$0.004 \\
			\midrule
		\end{tabular}
	}
\end{table}

\subsubsection{LDL algorithm Predictive Performance}
The reason to use the LE algorithm is that we need to recover the label distributions for LDL training. For the purpose of verifying the correctness and effectiveness of our proposed LE algorithm, we conducted an experiment to compare predictions depending on the recovered label distributions with those made by the LDL model trained on real label distributions. Moreover, for further evaluation of the proposed BD-LDL, we selected BD-LDL and SA-BFGS as the LDL model in this experiment. Owing to the page limit, we hereby present only the experimental results measured with Chebyshev. The prediction results achieved from SA-BFGS and BD-LDL are visualized in Figs. 1 and 2 in terms of histograms. Note that `Ground-truth' appearing in the figures represents the results depending on the real label distributions. We regard these results as a benchmark instead of taking them into consideration while conducting the evaluation. Meanwhile, we use `FCM', `KM', `LP', `ML', `GLLE' and `BD-LE' to represent the performance of the corresponding LE algorithm in this experiment. As illustrated in Figs. 1-2, although the prediction on datasets movie and SUB\_3DFE is worse than other datasets, `BD-LE' is still relatively close to `Ground-Truth' in most cases, especially in the first to eleventh datasets. We must mention that `BD-LE' combined with BD-LDL is closer to `Ground-truth' than with SA-BFGS over all cases. This indicates that such a reconstruction constraint is generalized enough to bring the improvement into both LE and LDL algorithms simultaneously.
\begin{figure}[]
	\centering
	\includegraphics[width=1\columnwidth]{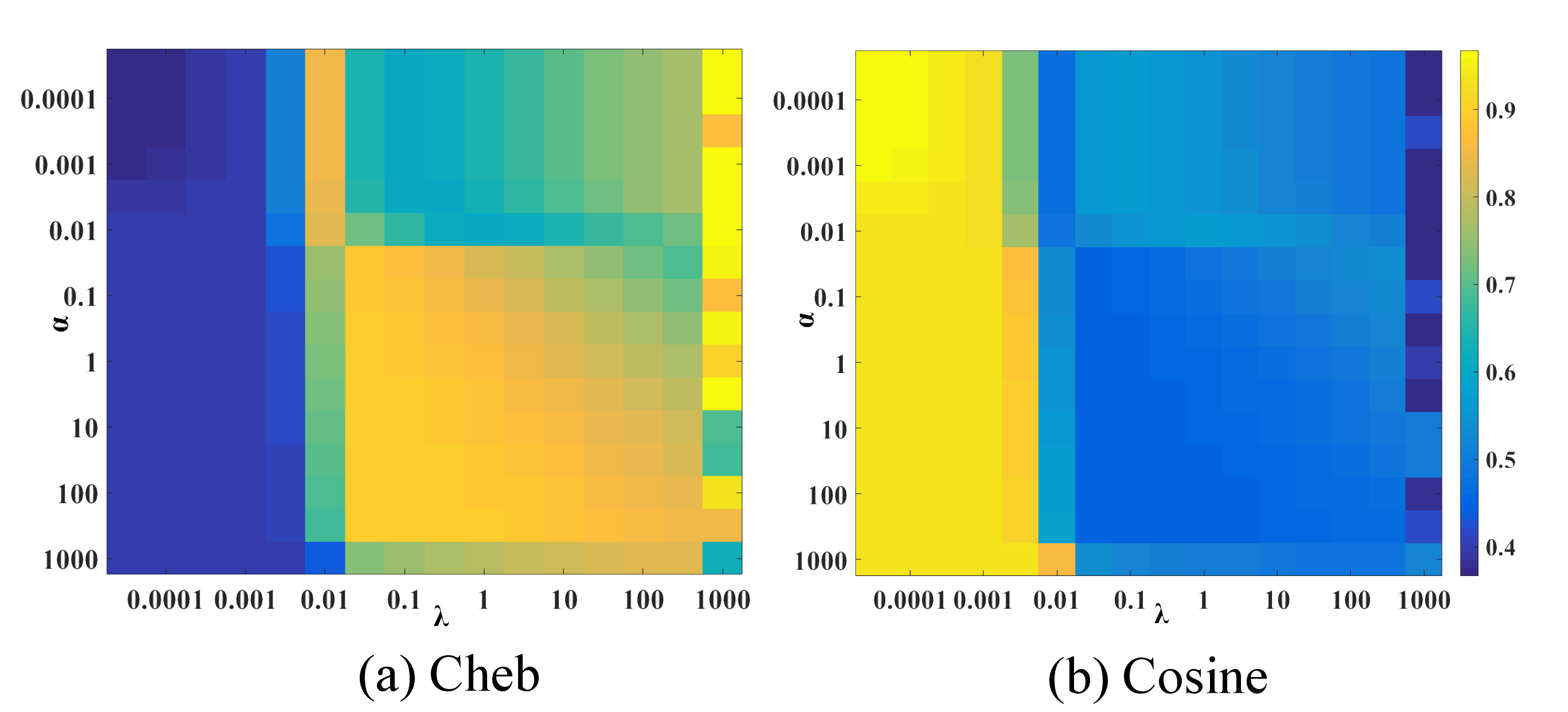}
	\caption{Influence of parameter $\lambda$ and $\alpha$ on dataset \textit{cold} in BD-LE}
	\label{fig:PARAMETER_LDL}
\end{figure}
\begin{figure}[]
	\centering
	\includegraphics[width=1\columnwidth]{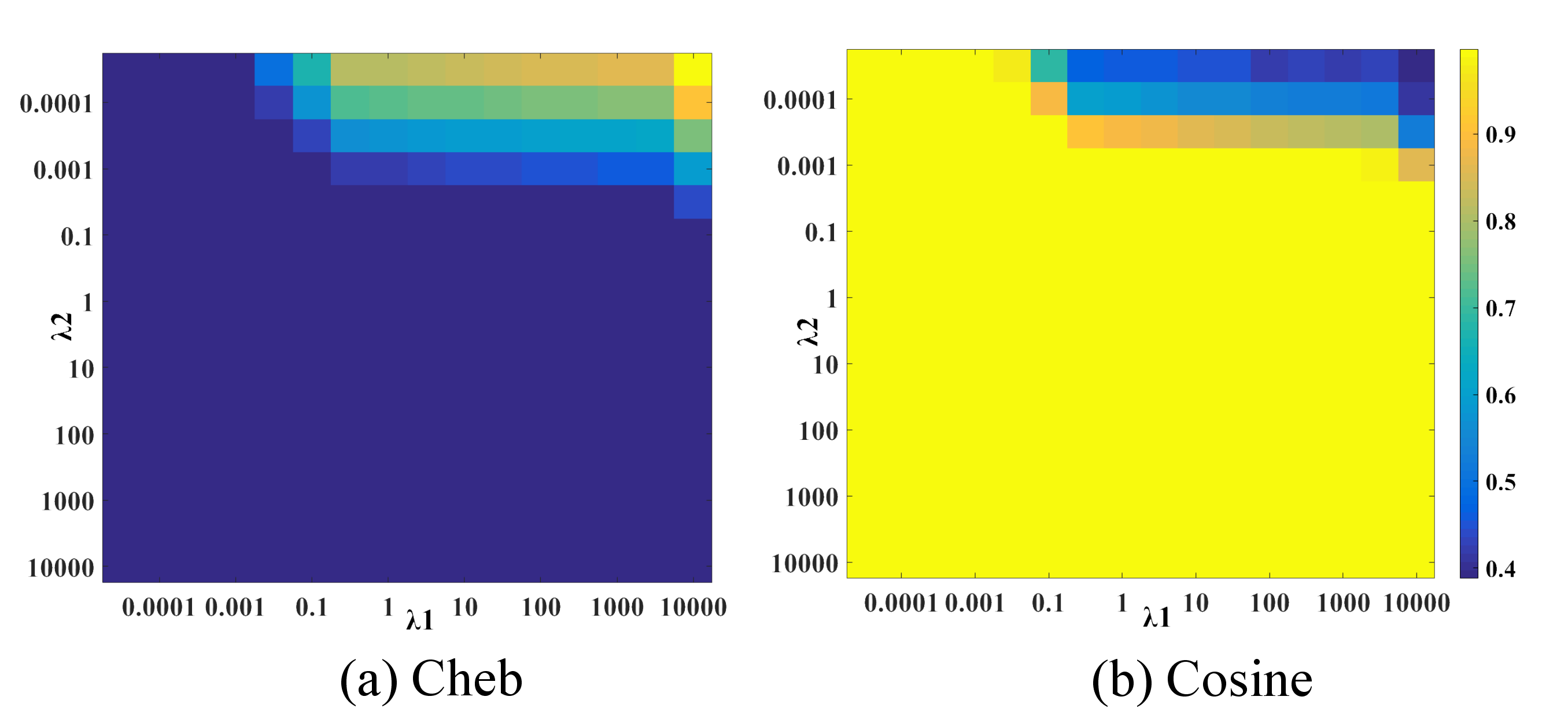}
	\caption{Influence of parameter $\lambda_{1}$ and $\lambda_{2}$ on dataset \textit{cold} in BD-LDL}
	\label{fig:PARAMETER_LE}
\end{figure} 

\subsection{Ablation Experiment}
It is clear that the bi-directional loss for either LE or LDL consists of two parts, i.e., naive mapping loss and the reconstruction loss. In order to further demonstrate the effectiveness of the additional reconstruction term, we conduct the ablation experiment measured with Canberra and Intersection on the whole 13 datasets. We call the unidirectional algorithm without reconstruction term as \textit{UD-LE} and \textit{UD-LE} respectively which are fomulated as:
\begin{equation}
\min _{W} L(\hat{W})+ \Omega(W)
\end{equation}
\begin{equation}
\min _{\theta} L(\theta, S)+\lambda \Omega(\theta, S)
\end{equation} 
Since the objective function of UD-LE is identical to that of GLLE, the corresponding comparison can be referred to Tables 2 and 3. The LDL prediction results in metrics of Canberra and Intersection are tabulated in Table 6.

From Table 6 we can see that BD-LDL gains the superior results in all benchmark datasets, i.e., introducing the reconstruction term can truly boost the performance of LDL algorithm. It is expected that the top-3 improvements are achieved in datasets Natural Scene, Movie, SBU\_3DFE respectively which are equiped with the relative large dimensional gap between the feature and label space.

\subsection{Influence of Parameters}
To examine the robustness of the proposed algorithms, we also analyze the influence of trade-off parameters in the experiments, including $\lambda_{1}$, $\lambda_{2}$ in BD-LDL as well as $\alpha$ and $\lambda$ in BD-LE. We run BD-LE with $\alpha$ and $\lambda$ whose value range is [$10^{-4}$,$10^{3}$], and parameters $\lambda_{1}$ and $\lambda_{2}$ involved in BD-LDL use the same candidate label set as well. Owing to the page limit, we only show in this paper the experimental results on Yeast-cold dataset which are measured with Chebyshev and Cosine. For further evaluation, the results are visualized with different colors in Figs. 5 and 6. When measured with Chebyshev, a smaller value means better performance and closer to blue; by contrast, with cosine, a larger value indicates better performance and closer to red.

It is clear from Fig. 5 that when $\lambda$ falls in a certain range [$10^{-4}$,$10^{-1}$], we can achieve relatively good recovery results with any values of $\alpha$. After conducting several experiments, we draw a conclusion for BD-LE, namely that when both $\alpha$ and $\lambda$ are about $10^{-2}$, the best performance is obtained. Concerning the parameters of BD-LDL, when the value of $\lambda_{1}$ is selected within the range [$10^{-1}$,$10^{3}$], the color varies in an extremely steady way, which means that the performance is not sensitive to this hyper parameter in that particular range. In addition, we can also see from Fig. 6 that $\lambda_{1}$ has a stronger influence on the performance than $\lambda_{2}$ when the value of $\lambda_{2}$ is within the range [$10^{-4}$,$10^{-1}$].

\section{Conclusion}
Previous studies have shown that the LDL method can effectively solve label ambiguity problems whereas the LE method is able to recover label distributions from logical labels. To improve the performance of LDL and LE methods, we propose a new loss function that combines the mapping error with the reconstruction error to leverage the missing information caused by the dimensional gap between the input space and the output one. Sufficient experiments have been conducted to show that the proposed loss function is sufficiently generalized for application in both LDL and LE with improvement. In the future, we will explore if there exists an end-to-end way to recover the label distributions with the supervision of LDL training process.

\bibliographystyle{cas-model2-names}
\bibliography{mm}

\begin{thebibliography}{43}
\expandafter\ifx\csname natexlab\endcsname\relax\def\natexlab#1{#1}\fi
\providecommand{\url}[1]{\texttt{#1}}
\providecommand{\href}[2]{#2}
\providecommand{\path}[1]{#1}
\providecommand{\DOIprefix}{doi:}
\providecommand{\ArXivprefix}{arXiv:}
\providecommand{\URLprefix}{URL: }
\providecommand{\Pubmedprefix}{pmid:}
\providecommand{\doi}[1]{\href{http://dx.doi.org/#1}{\path{#1}}}
\providecommand{\Pubmed}[1]{\href{pmid:#1}{\path{#1}}}
\providecommand{\bibinfo}[2]{#2}
\ifx\xfnm\relax \def\xfnm[#1]{\unskip,\space#1}\fi
\bibitem[{Berger et~al.(1996)Berger, Pietra and Pietra}]{berger1996maximum}
\bibinfo{author}{Berger, A.L.}, \bibinfo{author}{Pietra, V.J.D.},
  \bibinfo{author}{Pietra, S.A.D.}, \bibinfo{year}{1996}.
\newblock \bibinfo{title}{A maximum entropy approach to natural language
  processing}.
\newblock \bibinfo{journal}{Computational linguistics} \bibinfo{volume}{22},
  \bibinfo{pages}{39--71}.
\bibitem[{Boureau et~al.(2008)Boureau, Cun et~al.}]{boureau2008sparseFeature}
\bibinfo{author}{Boureau, Y.l.}, \bibinfo{author}{Cun, Y.L.}, et~al.,
  \bibinfo{year}{2008}.
\newblock \bibinfo{title}{Sparse feature learning for deep belief networks},
  in: \bibinfo{booktitle}{Advances in neural information processing systems},
  pp. \bibinfo{pages}{1185--1192}.
\bibitem[{Boyd et~al.(2011)Boyd, Parikh, Chu, Peleato, Eckstein
  et~al.}]{boyd2011distributedADMM}
\bibinfo{author}{Boyd, S.}, \bibinfo{author}{Parikh, N.}, \bibinfo{author}{Chu,
  E.}, \bibinfo{author}{Peleato, B.}, \bibinfo{author}{Eckstein, J.}, et~al.,
  \bibinfo{year}{2011}.
\newblock \bibinfo{title}{Distributed optimization and statistical learning via
  the alternating direction method of multipliers}.
\newblock \bibinfo{journal}{Foundations and Trends{\textregistered} in Machine
  learning} \bibinfo{volume}{3}, \bibinfo{pages}{1--122}.
\bibitem[{Cheng et~al.(2019)Cheng, Zhao, Wang and Pei}]{cheng2019multi}
\bibinfo{author}{Cheng, Y.}, \bibinfo{author}{Zhao, D.}, \bibinfo{author}{Wang,
  Y.}, \bibinfo{author}{Pei, G.}, \bibinfo{year}{2019}.
\newblock \bibinfo{title}{Multi-label learning with kernel extreme learning
  machine autoencoder}.
\newblock \bibinfo{journal}{Knowledge-Based Systems} \bibinfo{volume}{178},
  \bibinfo{pages}{1--10}.
\bibitem[{Eisen et~al.(1998)Eisen, Spellman, Brown and
  Botstein}]{eisen1998Genecluster}
\bibinfo{author}{Eisen, M.B.}, \bibinfo{author}{Spellman, P.T.},
  \bibinfo{author}{Brown, P.O.}, \bibinfo{author}{Botstein, D.},
  \bibinfo{year}{1998}.
\newblock \bibinfo{title}{Cluster analysis and display of genome-wide
  expression patterns}.
\newblock \bibinfo{journal}{Proceedings of the National Academy of Sciences}
  \bibinfo{volume}{95}, \bibinfo{pages}{14863--14868}.
\bibitem[{El~Gayar et~al.(2006)El~Gayar, Schwenker and Palm}]{el2006studyKNN}
\bibinfo{author}{El~Gayar, N.}, \bibinfo{author}{Schwenker, F.},
  \bibinfo{author}{Palm, G.}, \bibinfo{year}{2006}.
\newblock \bibinfo{title}{A study of the robustness of knn classifiers trained
  using soft labels}, in: \bibinfo{booktitle}{IAPR Workshop on Artificial
  Neural Networks in Pattern Recognition}, \bibinfo{organization}{Springer}.
  pp. \bibinfo{pages}{67--80}.
\bibitem[{Fan et~al.(2017)Fan, Liu, Li, Guo, Samal, Wan and
  Li}]{fan2017labelreNetFacial}
\bibinfo{author}{Fan, Y.Y.}, \bibinfo{author}{Liu, S.}, \bibinfo{author}{Li,
  B.}, \bibinfo{author}{Guo, Z.}, \bibinfo{author}{Samal, A.},
  \bibinfo{author}{Wan, J.}, \bibinfo{author}{Li, S.Z.}, \bibinfo{year}{2017}.
\newblock \bibinfo{title}{Label distribution-based facial attractiveness
  computation by deep residual learning}.
\newblock \bibinfo{journal}{IEEE Transactions on Multimedia}
  \bibinfo{volume}{20}, \bibinfo{pages}{2196--2208}.
\bibitem[{Geng(2016)}]{geng2016LDL}
\bibinfo{author}{Geng, X.}, \bibinfo{year}{2016}.
\newblock \bibinfo{title}{Label distribution learning}.
\newblock \bibinfo{journal}{IEEE Transactions on Knowledge and Data
  Engineering} \bibinfo{volume}{28}, \bibinfo{pages}{1734--1748}.
\bibitem[{Geng and Hou(2015)}]{geng2015preMovieOpnion}
\bibinfo{author}{Geng, X.}, \bibinfo{author}{Hou, P.}, \bibinfo{year}{2015}.
\newblock \bibinfo{title}{Pre-release prediction of crowd opinion on movies by
  label distribution learning}, in: \bibinfo{booktitle}{Twenty-Fourth
  International Joint Conference on Artificial Intelligence}.
\bibitem[{Geng and Ji(2013)}]{geng2013LDL}
\bibinfo{author}{Geng, X.}, \bibinfo{author}{Ji, R.}, \bibinfo{year}{2013}.
\newblock \bibinfo{title}{Label distribution learning}, in:
  \bibinfo{booktitle}{Proceedings of the 2013 IEEE 13th International
  Conference on Data Mining Workshops}, \bibinfo{organization}{IEEE Computer
  Society}. pp. \bibinfo{pages}{377--383}.
\bibitem[{Geng et~al.(2010)Geng, Smith-Miles and
  Zhou}]{geng2010facialestimation}
\bibinfo{author}{Geng, X.}, \bibinfo{author}{Smith-Miles, K.},
  \bibinfo{author}{Zhou, Z.H.}, \bibinfo{year}{2010}.
\newblock \bibinfo{title}{Facial age estimation by learning from label
  distributions}, in: \bibinfo{booktitle}{Proceedings of the Twenty-Fourth AAAI
  Conference on Artificial Intelligence}, \bibinfo{organization}{AAAI Press}.
  pp. \bibinfo{pages}{451--456}.
\bibitem[{Geng et~al.(2014)Geng, Wang and Xia}]{geng2014facialadaptive}
\bibinfo{author}{Geng, X.}, \bibinfo{author}{Wang, Q.}, \bibinfo{author}{Xia,
  Y.}, \bibinfo{year}{2014}.
\newblock \bibinfo{title}{Facial age estimation by adaptive label distribution
  learning}, in: \bibinfo{booktitle}{2014 22nd International Conference on
  Pattern Recognition}, \bibinfo{organization}{IEEE}. pp.
  \bibinfo{pages}{4465--4470}.
\bibitem[{Geng et~al.(2013)Geng, Yin and Zhou}]{geng2013facialestimation}
\bibinfo{author}{Geng, X.}, \bibinfo{author}{Yin, C.}, \bibinfo{author}{Zhou,
  Z.H.}, \bibinfo{year}{2013}.
\newblock \bibinfo{title}{Facial age estimation by learning from label
  distributions}.
\newblock \bibinfo{journal}{IEEE transactions on pattern analysis and machine
  intelligence} \bibinfo{volume}{35}, \bibinfo{pages}{2401--2412}.
\bibitem[{{Golub} and {Loan}(1996)}]{golub1996matrix}
\bibinfo{author}{{Golub}, G.H.}, \bibinfo{author}{{Loan}, C.F.V.},
  \bibinfo{year}{1996}.
\newblock \bibinfo{title}{Matrix computations (3rd ed.)} .
\bibitem[{He et~al.(2019)He, Yang, Gao, Liu and Yin}]{he2019joint}
\bibinfo{author}{He, Z.F.}, \bibinfo{author}{Yang, M.}, \bibinfo{author}{Gao,
  Y.}, \bibinfo{author}{Liu, H.D.}, \bibinfo{author}{Yin, Y.},
  \bibinfo{year}{2019}.
\newblock \bibinfo{title}{Joint multi-label classification and label
  correlations with missing labels and feature selection}.
\newblock \bibinfo{journal}{Knowledge-Based Systems} \bibinfo{volume}{163},
  \bibinfo{pages}{145--158}.
\bibitem[{Hou et~al.(2016)Hou, Geng and Zhang}]{hou2016manifold}
\bibinfo{author}{Hou, P.}, \bibinfo{author}{Geng, X.}, \bibinfo{author}{Zhang,
  M.L.}, \bibinfo{year}{2016}.
\newblock \bibinfo{title}{Multi-label manifold learning}, in:
  \bibinfo{booktitle}{Thirtieth AAAI Conference on Artificial Intelligence}.
\bibitem[{Jia et~al.(2018)Jia, Li, Liu and Zhang}]{jia2018labelCorrelation}
\bibinfo{author}{Jia, X.}, \bibinfo{author}{Li, W.}, \bibinfo{author}{Liu, J.},
  \bibinfo{author}{Zhang, Y.}, \bibinfo{year}{2018}.
\newblock \bibinfo{title}{Label distribution learning by exploiting label
  correlations}, in: \bibinfo{booktitle}{Thirty-Second AAAI Conference on
  Artificial Intelligence}.
\bibitem[{Jia et~al.(2019a)Jia, Ren, Chen, Wang, Zhu and Long}]{jia2019weakly}
\bibinfo{author}{Jia, X.}, \bibinfo{author}{Ren, T.}, \bibinfo{author}{Chen,
  L.}, \bibinfo{author}{Wang, J.}, \bibinfo{author}{Zhu, J.},
  \bibinfo{author}{Long, X.}, \bibinfo{year}{2019}a.
\newblock \bibinfo{title}{Weakly supervised label distribution learning based
  on transductive matrix completion with sample correlations}.
\newblock \bibinfo{journal}{Pattern Recognition Letters} \bibinfo{volume}{125},
  \bibinfo{pages}{453--462}.
\bibitem[{Jia et~al.(2019b)Jia, Zheng, Li, Zhang and Li}]{jia2019facialEDLLRL}
\bibinfo{author}{Jia, X.}, \bibinfo{author}{Zheng, X.}, \bibinfo{author}{Li,
  W.}, \bibinfo{author}{Zhang, C.}, \bibinfo{author}{Li, Z.},
  \bibinfo{year}{2019}b.
\newblock \bibinfo{title}{Facial emotion distribution learning by exploiting
  low-rank label correlations locally}, in: \bibinfo{booktitle}{Proceedings of
  the IEEE Conference on Computer Vision and Pattern Recognition}, pp.
  \bibinfo{pages}{9841--9850}.
\bibitem[{Jiang et~al.(2006)Jiang, Yi and Lv}]{jiang2006fuzzySVM}
\bibinfo{author}{Jiang, X.}, \bibinfo{author}{Yi, Z.}, \bibinfo{author}{Lv,
  J.C.}, \bibinfo{year}{2006}.
\newblock \bibinfo{title}{Fuzzy svm with a new fuzzy membership function}.
\newblock \bibinfo{journal}{Neural Computing \& Applications}
  \bibinfo{volume}{15}, \bibinfo{pages}{268--276}.
\bibitem[{Kodirov and Gong(2017)}]{kodirov2017SAE}
\bibinfo{author}{Kodirov, Elyor, X.T.}, \bibinfo{author}{Gong, S.},
  \bibinfo{year}{2017}.
\newblock \bibinfo{title}{Semantic autoencoder for zero-shot learning}, in:
  \bibinfo{booktitle}{Proceedings of the IEEE Conference on Computer Vision and
  Pattern Recognition}, pp. \bibinfo{pages}{3174--3183}.
\bibitem[{Lyons et~al.(1998)Lyons, Akamatsu, Kamachi and
  Gyoba}]{lyons1998coding-facial}
\bibinfo{author}{Lyons, M.}, \bibinfo{author}{Akamatsu, S.},
  \bibinfo{author}{Kamachi, M.}, \bibinfo{author}{Gyoba, J.},
  \bibinfo{year}{1998}.
\newblock \bibinfo{title}{Coding facial expressions with gabor wavelets}, in:
  \bibinfo{booktitle}{Proceedings Third IEEE international conference on
  automatic face and gesture recognition}, \bibinfo{organization}{IEEE}. pp.
  \bibinfo{pages}{200--205}.
\bibitem[{Ma et~al.(2017)Ma, Tian, Zhang and Chow}]{ma2017multi}
\bibinfo{author}{Ma, J.}, \bibinfo{author}{Tian, Z.}, \bibinfo{author}{Zhang,
  H.}, \bibinfo{author}{Chow, T.W.}, \bibinfo{year}{2017}.
\newblock \bibinfo{title}{Multi-label low-dimensional embedding with missing
  labels}.
\newblock \bibinfo{journal}{Knowledge-Based Systems} \bibinfo{volume}{137},
  \bibinfo{pages}{65--82}.
\bibitem[{Malouf(2002)}]{Maxinum_entropy}
\bibinfo{author}{Malouf, R.}, \bibinfo{year}{2002}.
\newblock \bibinfo{title}{A comparison of algorithms for maximum entropy
  parameter estimation}, in: \bibinfo{booktitle}{proceedings of the 6th
  conference on Natural language learning-Volume 20},
  \bibinfo{organization}{Association for Computational Linguistics}. pp.
  \bibinfo{pages}{1--7}.
\bibitem[{Melin and Castillo(2005)}]{melin2005hybrid}
\bibinfo{author}{Melin, P.}, \bibinfo{author}{Castillo, O.},
  \bibinfo{year}{2005}.
\newblock \bibinfo{title}{Hybrid intelligent systems for pattern recognition
  using soft computing: An evolutionary approach for neural networks and fuzzy
  systems}. volume \bibinfo{volume}{172}.
\newblock \bibinfo{publisher}{Springer Science \& Business Media}.
\bibitem[{Nocedal and Wright(2006)}]{nocedal2006numericalOPtimization}
\bibinfo{author}{Nocedal, J.}, \bibinfo{author}{Wright, S.},
  \bibinfo{year}{2006}.
\newblock \bibinfo{title}{Numerical optimization}.
\newblock \bibinfo{publisher}{Springer Science \& Business Media}.
\bibitem[{Ren et~al.(2019a)Ren, Jia, Li, Chen and Li}]{ren2019specific}
\bibinfo{author}{Ren, T.}, \bibinfo{author}{Jia, X.}, \bibinfo{author}{Li, W.},
  \bibinfo{author}{Chen, L.}, \bibinfo{author}{Li, Z.}, \bibinfo{year}{2019}a.
\newblock \bibinfo{title}{Label distribution learning with label-specific
  features.}, in: \bibinfo{booktitle}{Proceedings of the 28th International
  Joint Conference on Artificial Intelligence}, pp.
  \bibinfo{pages}{3318--3324}.
\bibitem[{Ren et~al.(2019b)Ren, Jia, Li and Zhao}]{ren2019label}
\bibinfo{author}{Ren, T.}, \bibinfo{author}{Jia, X.}, \bibinfo{author}{Li, W.},
  \bibinfo{author}{Zhao, S.}, \bibinfo{year}{2019}b.
\newblock \bibinfo{title}{Label distribution learning with label correlations
  via low-rank approximation}, in: \bibinfo{booktitle}{Proceedings of the 28th
  International Joint Conference on Artificial Intelligence},
  \bibinfo{organization}{AAAI Press}. pp. \bibinfo{pages}{3325--3331}.
\bibitem[{Tsoumakas and Katakis(2007)}]{tsoumakas2007multiMLL}
\bibinfo{author}{Tsoumakas, G.}, \bibinfo{author}{Katakis, I.},
  \bibinfo{year}{2007}.
\newblock \bibinfo{title}{Multi-label classification: An overview}.
\newblock \bibinfo{journal}{International Journal of Data Warehousing and
  Mining (IJDWM)} \bibinfo{volume}{3}, \bibinfo{pages}{1--13}.
\bibitem[{Wang and Zhang(2007)}]{wang2007labelpropagation}
\bibinfo{author}{Wang, F.}, \bibinfo{author}{Zhang, C.}, \bibinfo{year}{2007}.
\newblock \bibinfo{title}{Label propagation through linear neighborhoods}.
\newblock \bibinfo{journal}{IEEE Transactions on Knowledge and Data
  Engineering} \bibinfo{volume}{20}, \bibinfo{pages}{55--67}.
\bibitem[{Xing et~al.(2016)Xing, Geng and Xue}]{xing2016logisticBoosting}
\bibinfo{author}{Xing, C.}, \bibinfo{author}{Geng, X.}, \bibinfo{author}{Xue,
  H.}, \bibinfo{year}{2016}.
\newblock \bibinfo{title}{Logistic boosting regression for label distribution
  learning}, in: \bibinfo{booktitle}{Proceedings of the IEEE conference on
  computer vision and pattern recognition}, pp. \bibinfo{pages}{4489--4497}.
\bibitem[{Xu and Zhou(2017)}]{xu2017incompleteLDL}
\bibinfo{author}{Xu, M.}, \bibinfo{author}{Zhou, Z.H.}, \bibinfo{year}{2017}.
\newblock \bibinfo{title}{Incomplete label distribution learning.}, in:
  \bibinfo{booktitle}{IJCAI}, pp. \bibinfo{pages}{3175--3181}.
\bibitem[{Xu et~al.(2019a)Xu, Liu and Geng}]{xu2019labelTKDE}
\bibinfo{author}{Xu, N.}, \bibinfo{author}{Liu, Y.P.}, \bibinfo{author}{Geng,
  X.}, \bibinfo{year}{2019}a.
\newblock \bibinfo{title}{Label enhancement for label distribution learning}.
\newblock \bibinfo{journal}{IEEE Transactions on Knowledge and Data
  Engineering} .
\bibitem[{Xu et~al.(2019b)Xu, Lv and Geng}]{XuLv-14}
\bibinfo{author}{Xu, N.}, \bibinfo{author}{Lv, J.}, \bibinfo{author}{Geng, X.},
  \bibinfo{year}{2019}b.
\newblock \bibinfo{title}{Partial label learning via label enhancement}.
\bibitem[{Xu et~al.(2018)Xu, Tao and Geng}]{xu2018LE}
\bibinfo{author}{Xu, N.}, \bibinfo{author}{Tao, A.}, \bibinfo{author}{Geng,
  X.}, \bibinfo{year}{2018}.
\newblock \bibinfo{title}{Label enhancement for label distribution learning.},
  in: \bibinfo{booktitle}{IJCAI}, pp. \bibinfo{pages}{2926--2932}.
\bibitem[{Xu et~al.(2019c)Xu, Shang and Shen}]{xu2019latent}
\bibinfo{author}{Xu, S.}, \bibinfo{author}{Shang, L.}, \bibinfo{author}{Shen,
  F.}, \bibinfo{year}{2019}c.
\newblock \bibinfo{title}{Latent semantics encoding for label distribution
  learning.}, in: \bibinfo{booktitle}{Proceedings of the 28th International
  Joint Conference on Artificial Intelligence}, pp.
  \bibinfo{pages}{3982--3988}.
\bibitem[{Xu et~al.(2016)Xu, Yang, Yu, Yu, Yang and Tsang}]{xu2016multi}
\bibinfo{author}{Xu, S.}, \bibinfo{author}{Yang, X.}, \bibinfo{author}{Yu, H.},
  \bibinfo{author}{Yu, D.J.}, \bibinfo{author}{Yang, J.},
  \bibinfo{author}{Tsang, E.C.}, \bibinfo{year}{2016}.
\newblock \bibinfo{title}{Multi-label learning with label-specific feature
  reduction}.
\newblock \bibinfo{journal}{Knowledge-Based Systems} \bibinfo{volume}{104},
  \bibinfo{pages}{52--61}.
\bibitem[{Yuan(1991)}]{yuan1991modifiedBFGS}
\bibinfo{author}{Yuan, Y.x.}, \bibinfo{year}{1991}.
\newblock \bibinfo{title}{A modified bfgs algorithm for unconstrained
  optimization}.
\newblock \bibinfo{journal}{IMA Journal of Numerical Analysis}
  \bibinfo{volume}{11}, \bibinfo{pages}{325--332}.
\bibitem[{Zheng et~al.(2018)Zheng, Jia and
  Li}]{zheng2018labelCorrelationSample}
\bibinfo{author}{Zheng, X.}, \bibinfo{author}{Jia, X.}, \bibinfo{author}{Li,
  W.}, \bibinfo{year}{2018}.
\newblock \bibinfo{title}{Label distribution learning by exploiting sample
  correlations locally}, in: \bibinfo{booktitle}{Thirty-Second AAAI Conference
  on Artificial Intelligence}.
\bibitem[{Zhou et~al.(2015)Zhou, Xue and Geng}]{zhou2015emotion}
\bibinfo{author}{Zhou, Y.}, \bibinfo{author}{Xue, H.}, \bibinfo{author}{Geng,
  X.}, \bibinfo{year}{2015}.
\newblock \bibinfo{title}{Emotion distribution recognition from facial
  expressions}, in: \bibinfo{booktitle}{Proceedings of the 23rd ACM
  international conference on Multimedia}, \bibinfo{organization}{ACM}. pp.
  \bibinfo{pages}{1247--1250}.
\bibitem[{Zhou et~al.(2012)Zhou, Zhang, Huang and Li}]{zhou2012multi}
\bibinfo{author}{Zhou, Z.H.}, \bibinfo{author}{Zhang, M.L.},
  \bibinfo{author}{Huang, S.J.}, \bibinfo{author}{Li, Y.F.},
  \bibinfo{year}{2012}.
\newblock \bibinfo{title}{Multi-instance multi-label learning}.
\newblock \bibinfo{journal}{Artificial Intelligence} \bibinfo{volume}{176},
  \bibinfo{pages}{2291--2320}.
\bibitem[{Zhu et~al.(2005)Zhu, Lafferty and Rosenfeld}]{zhu2005semiGraph}
\bibinfo{author}{Zhu, X.}, \bibinfo{author}{Lafferty, J.},
  \bibinfo{author}{Rosenfeld, R.}, \bibinfo{year}{2005}.
\newblock \bibinfo{title}{Semi-supervised learning with graphs}.
\newblock Ph.D. thesis. Carnegie Mellon University, language technologies
  institute, school of~….
\bibitem[{{Zhu} et~al.(2017){Zhu}, {Suk}, {Wang}, {Lee} and {Shen}}]{zhu2017a}
\bibinfo{author}{{Zhu}, X.}, \bibinfo{author}{{Suk}, H.I.},
  \bibinfo{author}{{Wang}, L.}, \bibinfo{author}{{Lee}, S.W.},
  \bibinfo{author}{{Shen}, D.}, \bibinfo{year}{2017}.
\newblock \bibinfo{title}{A novel relational regularization feature selection
  method for joint regression and classification in ad diagnosis.}
\newblock \bibinfo{journal}{Medical Image Analysis} \bibinfo{volume}{38},
  \bibinfo{pages}{205--214}.

\end{thebibliography}

\end{document}